\documentclass[table]{bmvc2k}
\usepackage{amsthm,amsmath,amssymb}
\usepackage{mathrsfs}
\usepackage{multirow}
\usepackage{color}
\usepackage{enumitem}
\usepackage{pifont}
\newcommand{\cmark}{\ding{51}}
\usepackage{caption}
\usepackage{graphicx}
\usepackage{booktabs}

\title{Transferring Domain-Agnostic Knowledge in Video Question Answering}

\addauthor{Tianran Wu}{tianran.wu@is.ids.osaka-u.ac.jp}{1}
\addauthor{Noa Garcia}{noagarcia@ids.osaka-u.ac.jp}{1}
\addauthor{Mayu Otani}{otani_mayu@cyberagentco.jp}{2}
\addauthor{Chenhui Chu}{chu@i.kyoto-u.ac.jp}{3}
\addauthor{Yuta Nakashima}{n_yuta@ids.osaka-u.ac.jp}{1}
\addauthor{Haruo Takemura}{takemura@cmc.osaka-u.ac.jp}{1}

\addinstitution{
 Osaka University\\
 Osaka, Japan
}
\addinstitution{
 CyberAgent, Inc.\\
 Japan
}
\addinstitution{
 Kyoto University\\
 Kyoto, Japan
}

\runninghead{Wu \etal}{TRANSFERRING KNOWLEDGE IN VIDEO QUESTION ANSWERING}

\def\ie{\emph{i.e}\bmvaOneDot}
\def\eg{\emph{e.g}\bmvaOneDot}

\def\etal{\emph{et al}\bmvaOneDot}

\begin{document}

\maketitle

\begin{abstract}
    Video question answering (VideoQA) is designed to answer a given question based on a relevant video clip. The current available large-scale datasets have made it possible to formulate VideoQA as the joint understanding of visual and language information. However, this training procedure is costly and still less competent with human performance. In this paper, we investigate a transfer learning method by the introduction of domain-agnostic knowledge and domain-specific knowledge. First, we develop a novel transfer learning framework, which finetunes the pre-trained model by applying domain-agnostic knowledge as the medium. Second, we construct a new VideoQA dataset with $21,412$ human-generated question-answer samples for comparable transfer of knowledge. Our experiments show that: (i) domain-agnostic knowledge is transferable and (ii) our proposed transfer learning framework can boost VideoQA performance effectively.
\end{abstract}

\section{Introduction}
\label{sec:intro}

Video question answering (VideoQA), which is designed to predict an answer to a given question based on a relevant video clip, requires comprehending both visual and linguistic content. This has made VideoQA an ideal testbed to evaluate current machine learning models, and researchers have made great efforts to advance the field. For example, TVQA~\cite{lei2018tvqa} presents a large-scale dataset and a model that leverages Faster-RCNN~\cite{ren2016faster} and LSTMs~\cite{hochreiter1997long} to process visual and language inputs, and the use of attention mechanisms~\cite{vaswani2017attention} has also achieved a great success~\cite{xue2017unifying,ye2017video,zhao2017video}. Recently, a new research direction in VideoQA has emerged, \ie, external knowledge-based VideoQA~\cite{garcia2020knowit,garcia2020knowledge}, which requires information that cannot be directly obtained from the videos or the question-answer (QA) pairs, and thus, cannot be learned from the dataset. In this task, therefore, a model needs to refer to knowledge from external sources. This configuration may be closer to real-world question answering. 

Knowledge for VideoQA, including external knowledge-based one, can be obtained in different ways: some knowledge can be learned from the videos and QA pairs in the dataset, and other knowledge can be extracted from a (external) knowledge base. Existing approaches use a dedicated set of sentences, each of which is associated with its respective QA pair \cite{garcia2020knowit} or a block of text found in the Internet \cite{garcia2020knowledge,marino2019ok}, as knowledge base in a specific domain, which are retrieved by the model during inference. This means that the knowledge base can be potentially replaced without re-training. In this case, a natural question arises here: \textit{How much is a trained VideoQA model's knowledge generalised for different domains?}

\begin{figure*}
\begin{center}
\includegraphics[width=12cm]{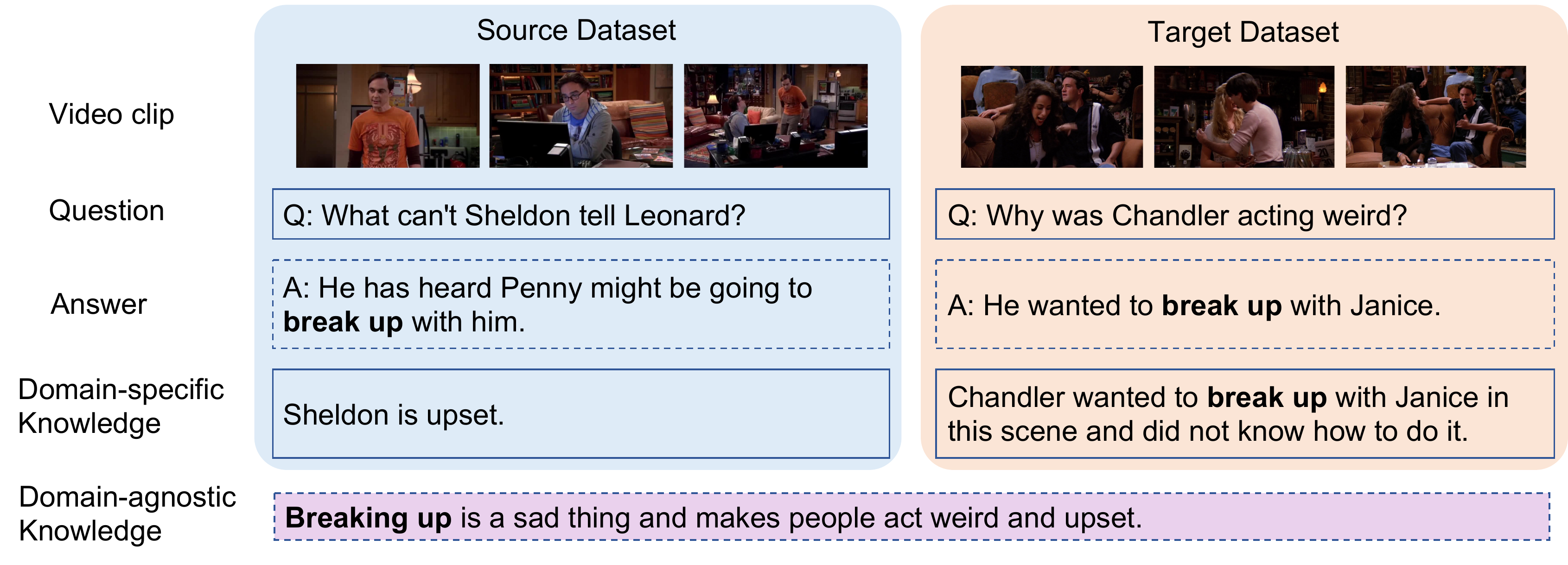}
\vspace{10pt}
\caption{Examples of transfer learning on different domains of VideoQA. The two datasets (source and target) have different domain-specific knowledge but share the same domain-agnostic knowledge. We argue that the latter can be transferred across datasets.}
\label{fig:overview}
\end{center}
\end{figure*}

In this paper, we study the transferability of knowledge learned in VideoQA. Specifically, we split the knowledge into two categories (Figure \ref{fig:overview}): \textit{domain-specific} and \textit{domain-agnostic}. The domain-specific knowledge is related to a particular domain of videos in a VideoQA dataset. This type of knowledge is acquired for each domain via training. On the other hand, the domain-agnostic knowledge involves (i) the so-called tacit knowledge, such as how to understand videos and questions as well as how to retrieve the desired piece of knowledge from the knowledge base and (ii) the so-called explicit knowledge that is still valid for different domains, such as common-sense. 

Based on this distinction, we propose a knowledge-oriented transfer learning for better generalisation of VideoQA models. We study the influence of knowledge across datasets and mitigate the inter-dataset information gap by effectively transferring the domain-agnostic knowledge. Our main contributions are summarised as follows:

\begin{itemize}[noitemsep,topsep=0pt]
  \item [1)]
  We show that, like humans, domain-specific and domain-agnostic knowledge also exists in VideoQA. Also, the domain-agnostic knowledge is transferable and can bring considerable performance improvement.
  \item [2)] 
  For transfer learning experiments on VideoQA, we construct a new dataset, KnowIT-X, containing $21,412$ QA pairs, each of them annotated with knowledge.\footnote{Available at \url{https://knowit-vqa.github.io/}} The video clips are sampled from a TV show, \textit{Friends},  which is a different domain from the standard model~\cite{garcia2020knowit}'s with \textit{The Big Bang Theory}.
\end{itemize}

\section{Related Work}
\label{sec:related}

\paragraph{Video question answering}

Traditionally, temporal visual information and language are processed with deep neural network (DNN)-based models and fused to predict the correct answer. For example, LSTMs~\cite{hochreiter1997long} are used to embed the temporal visual and textual features in \cite{lei2018tvqa}, and attention mechanisms \cite{vaswani2017attention}, which allows to focus only on specific parts of the input, bring significant improvements in \cite{xue2017unifying,ye2017video,zhao2017video,engin2021hidden}. Other work \cite{yang2020bert,yang2021comparative} applies Transformers~\cite{devlin2018bert} to capture the information from videos. For better fusion of independent visual and textual sources, Hirota \etal~\cite{hirota2021picture,hirota2021visual} propose to use the textual representations to understand the visual sources. Besides the independent processing of vision and language, other approaches~\cite{chadha2020iperceive, huang2020location} model the relationships between objects. Recent works like ~\cite{Yang_2021_ICCV} also focus on the VideoQA without much human annotation.

Another direction is to retrieve external information from knowledge bases (KB), which successfully extends the linguistic features and steers the model to a specific part of the visual content. For example, in visual question answering (VQA), a structured KB like ConceptNet \cite{speer2017conceptnet} or Freebase \cite{bollacker2008freebase} is adopted as extra inputs in \cite{wang2017fvqa,wu2016ask,wang2015explicit} and a generic unstructured KB is investigated in \cite{marino2019ok}. For VideaQA, KnowIT VQA~\cite{garcia2020knowit} is the first unstructured video-based dataset built by humans. ROLL \cite{garcia2020knowledge} leverages online knowledge to answer questions about video stories, showing the great potential of knowledge-based models in VideoQA.

\vspace{-14pt}
\paragraph{Transfer learning in vision and language}

Transfer learning has been rarely investigated in VideoQA, but it has been addressed in the related VQA tasks. Hu \etal~\cite{hu2018learning} study transfer learning between open-ended VQA datasets by a probabilistic model. Xu \etal~\cite{xu2019open} learn and make use of joint features across different modalities. Chao \etal~\cite{chao2018cross} propose a domain adaptation method to deal with the cross-dataset mismatch on images, questions, or answers. However, these approaches are not directly applicable to VideoQA due to the much richer information contained in videos. In addition, more emphasis has been put on processing the dataset itself, rather than studying the knowledge learnt by the model, which gives a more general solution to diverse domains of datasets. Therefore, this paper proposes a novel knowledge-oriented transfer learning method for VideoQA. To the extent of our knowledge, this is the first work that studies the transfer of knowledge in VideoQA.

\begin{figure*}
\begin{center}
\includegraphics[width=0.85\linewidth]{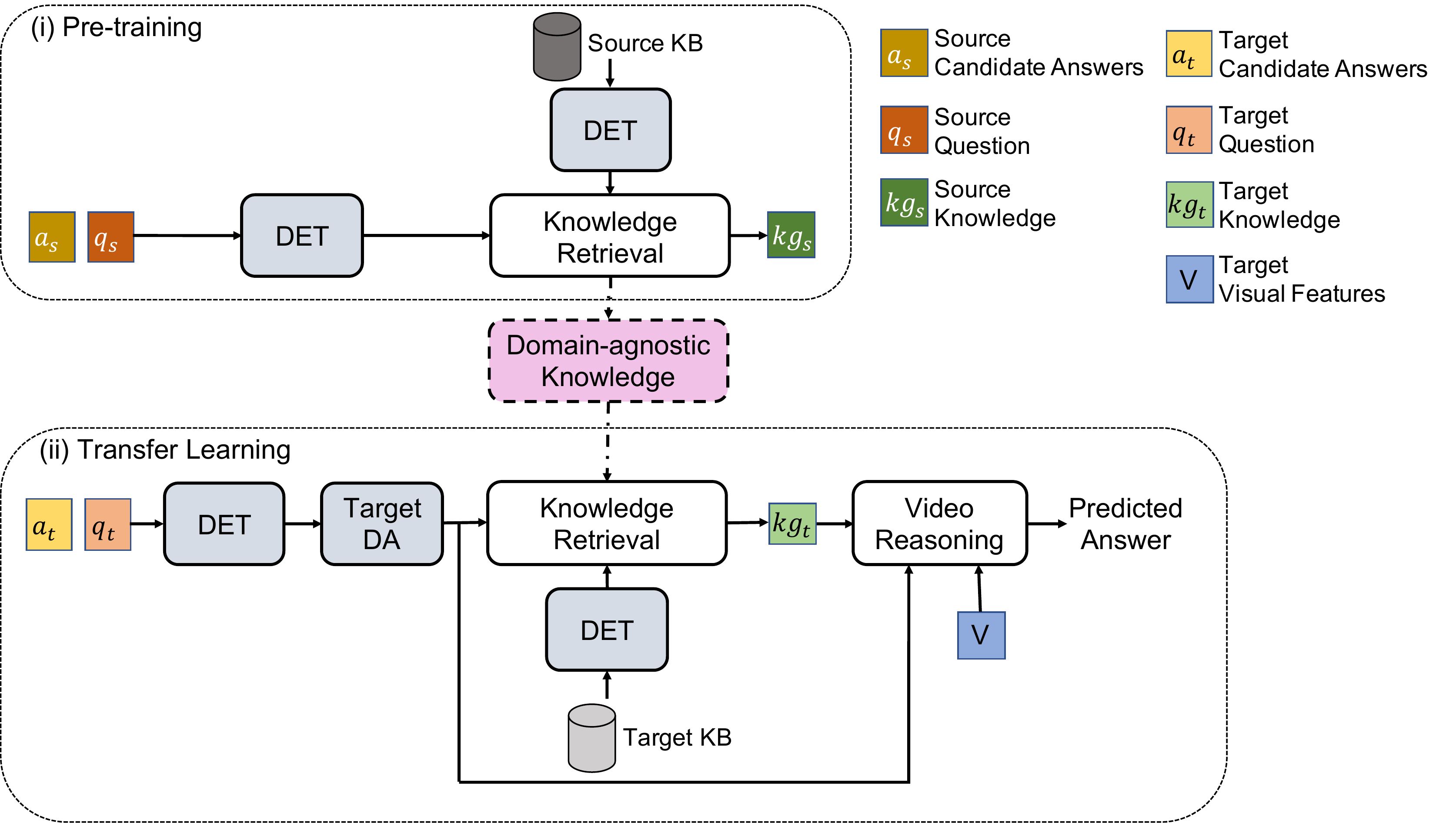}
\vspace{10pt}
\caption{Overview of our transfer learning method for knowledge-based VideoQA. In (i) pre-training, the knowledge retrieval module acquires domain-agnostic knowledge. In (ii) transfer learning, this knowledge is leveraged to improve the retrieval on the target dataset.}
\label{fig:method-overview}
\end{center}
\end{figure*}

\section{Methodology}
\label{sec:method}

Figure \ref{fig:method-overview} outlines the training pipeline of our proposed transfer learning method. First, in the \textit{pre-training} stage, we use a source dataset to train a knowledge retrieval module of a standard knowledge-based VideoQA model (Section \ref{sec:method-backbone}). In the knowledge retrieval module, a question and its candidate answers are used to retrieve the most relevant piece of knowledge from a KB. We design the pre-training stage so that the knowledge retrieval module learns to mitigate the domain-specific knowledge and focus on the domain-agnostic knowledge by pre-processing the source dataset with a domain-specific entity tagger~(DET) (Section \ref{sec:method-DET}). Then, in the \textit{transfer learning} stage, we pre-process the target dataset with DET and incorporate a data augmentation module (DA) (Section \ref{sec:method-dataaug}). Using the domain-agnostic knowledge acquired in the pre-training stage as a medium, we finetune the pre-trained knowledge retrieval module with the pre-processed target samples (Section \ref{sec:method-transfer}). Finally, the output of the knowledge retrieval module, together with the visual features, question, and candidate answers, is used in a video reasoning module to predict the correct answer (Section \ref{sec:method-prediction}). 

\subsection{Knowledge-based VideoQA model}
\label{sec:method-backbone}

We use ROCK \cite{garcia2020knowit} as the backbone VideoQA model. ROCK differs from standard VideoQA models~\cite{lei2018tvqa,chowdhury2018hierarchical,huang2020location} in that it explicitly uses external knowledge obtained from a KB, making it easier to study the problem of transferring knowledge across different domains.
ROCK takes a video clip, a question, and a set of candidate answers as input, and 1) retrieves relevant knowledge to answer the question from a KB and 2) predicts the correct answer by leveraging the input and the retrieved knowledge. Specifically, ROCK presents a three-stage pipeline, including (a) KB construction, (b) knowledge retrieval, and (c) video reasoning. The KB is constructed with annotated knowledge from the original dataset \cite{garcia2020knowit}. The knowledge retrieval module is a BERT network~\cite{devlin2018bert} trained to match the most relevant knowledge in the KB given a question and candidate answers. Finally, the video reasoning module fuses the retrieved knowledge with the rest of the language and visual inputs to predict the correct answer. For further details, we refer curious readers to \cite{garcia2020knowit}.

\subsection{Domain-specific entity tagger (DET)}
\label{sec:method-DET}
VideoQA models usually have strong in-domain generalisation across two sets of QA pairs with identical distributions \cite{lei2018tvqa}. However, in a transfer learning setup, a model is pre-trained on a source dataset and evaluated on a target dataset, where these datasets are from different origins and present different distributions. This leads to a large gap between the knowledge contained in the two datasets in terms of scenes, plots, and vocabulary. Thus, the domain-specific knowledge is specialised for a certain dataset (or domain) and may hinder the  transfer to out-of-domain data~\cite{noh2019transfer}. 
To mitigate the inter-domain difference, especially in the language modality, and  facilitate more effective learning of domain-agnostic knowledge, we propose a domain-specific entity tagger (DET) to recognise and tag the domain-specific knowledge with the named entity \cite{qi2020stanza} they belong to. As the entity names are more general expressions, we relieve the influence brought by the domain-specific knowledge in the source dataset so that the model does not need to know them. In this way, the transfer of domain-agnostic knowledge will be much easier by recognising the domain-specific knowledge.

Specifically, for a question $q$ in the dataset, such as `\textit{Why was Chandler acting weird?}', we recognise and tag the entity \textit{Chandler} as \verb'Person'. To maintain the original semantics and the grammatical structure, the tagged entity (\eg, \textit{Chandler}) and its entity type (\eg, \verb'Person') are inserted back into the original sentence in an appositive form. The question $\text{DET}(q)$ after this entity tagging becomes: `\textit{Why was Chandler, \underline{a person}, acting weird?}'.

We apply DET to question $q$, candidate answers $\{a_i\}_i$, and knowledge instance $k$ in both the source and the target dataset. We denote the set after applying this by $\text{DET}(\mathcal{X})$, \ie,
\begin{equation}
    \text{DET}(\mathcal{X}) = \{(\text{DET}(q), \{\text{DET}(a_i)\}_i, \text{DET}(k) \mid (q, \{a_i\}_i, k) \in \mathcal{X}\},
\end{equation}
where $\mathcal{X}$ denotes any VideoQA set.

\subsection{Target data augmentation (DA)}
\label{sec:method-dataaug}
When applying transfer learning from a source to a target dataset, the latter may not contain enough samples to learn sufficient domain-specific knowledge in the target domain, which may cause overfitting. To address this problem, we augment the training set $\mathcal T$ in the target dataset before transferring the pre-trained model. To augment the training set we propose to apply back translation~\cite{DBLP:conf/acl/SennrichHB16,hirota2021picture}, which translates a natural language sentence into a pivot language with the function ${\rm Translate}(\cdot)$, and then translates it back to the original language with the function ${\rm BackTranslate}(\cdot)$. With this technique, we incorporate twice as many samples as the original dataset, with a different structure but maintaining the original semantics. Also, for extremely small-scale datasets, back translation can be applied several times using multiple pivot languages. 

Formally, for any text $s$ (\ie, a question, an candidate answer, or a knowledge instance) in the training set,
\begin{equation}
\begin{aligned}
    s_{\rm PV} &= {\rm Translate}(s) \\
    s' &= {\rm BackTranslate}(s_{\rm PV}),
\end{aligned}
\end{equation}
where $s_{\rm PV}$ is the sample in the pivot language and $s'$ the augmented sample.
As some $s'_{t}$ can be almost the same as its original ($s'_t \approx s_t$), we identify them with a pre-trained BERT network \cite{devlin2018bert} and similarity threshold $\alpha$ to remove them. We incorporate the new $s'_{t}$ to the target training set $\mathcal{T}$, creating $\text{DA}(\mathcal{T}) = \mathcal{T} \cup \mathcal T_{\rm BK}$, where $\mathcal{T}_\textrm{BK}$ is the set of back-translated triplets of a question, a set of candidate answers, and a knowledge instance, after removal. We apply DA to all the questions and candidate answers in the target training set.

\subsection{Domain-agnostic knowledge transfer for knowledge retrieval}
\label{sec:method-transfer}

With the samples tagged with domain-specific knowledge and augmented by back translation, we next introduce the transfer of the domain-agnostic knowledge. Our VideoQA model is pre-trained on the training set $\mathcal S$ of the source dataset and then transferred to the target dataset. During pre-training, the knowledge retrieval module is trained to locate the most relevant knowledge instances $kg$ in the source KB given the input source question $q$ and source candidate answers $\{a$\}. Using $p_\theta(kg \mid q, \{a\})$, which denotes the probability of $kg$ being relevant to $q$ and $\{a\}$ parameterized by $\theta$, this can be formulated as a maximisation problem of the expected log-likelihood:
\begin{equation}
    \theta_{\rm Pre} =  \mathop{\arg\max}_{\theta}  \mathbb{E} [\log (p_{\theta}(kg \mid q, \{a\}))],
    \label{eq:training}
\end{equation}
where the expectation is computed over all $(q, \{a\}, kg) \in \text{DET}(\mathcal{S})$.
In the transfer learning stage, the pre-trained knowledge retrieval module is finetuned to find parameter $\theta_\text{FT}$ with the same training strategy as Eq.~(\ref{eq:training}) but the expectation is computed over all $(q, \{a\}, kg) \in \text{DA}(\text{DET}(\mathcal{T}))$ and the parameters are initialised by $\theta_\text{Pre}$.

\subsection{Answer prediction on the target domain}
\label{sec:method-prediction}
To predict the final answer, we extract visual features $v$ from the input video clips in the target dataset and fuse them in the video reasoning module together with question $q$, candidate answers $\{a_i\}_i$, and retrieved knowledge $\hat{kg} = \mathop{\arg\max}_{kg} p_{\theta_\text{FT}}(kg \mid q, \{a\})$ by the transferred knowledge retrieval module. We extract $v$ at three different scales of information: image features from video frames using ResNet50~\cite{he2016deep}, facial features with recogniser~\cite{parkhi2015deep}, and captions generated with \cite{xu2015show}. The textual data (\ie, $q$, $\{a_i\}_i$, $kg$) is concatenated and encoded in single vector $u_i$ per candidate answer for $i = 1, \dots, N_a$ ($N_a$ is the number of candidate answers), as in \cite{garcia2020knowit}. The visual and language features are concatenated and projected with a fully-connected layer to obtain an answer score. The predicted answer is the one with the highest score. The video reasoning module is trained with a multi-class cross-entropy loss.

\section{KnowIT-X VQA Dataset}
\label{sec:dataset}
We use KnowIT VQA \cite{garcia2020knowit} as source dataset, as it contains annotated knowledge for each sample in VideoQA. As there are no other comparable datasets with knowledge for VideoQA to use as target, we collect a new one following KnowIT VQA framework. We name it KnowIT-X VQA (X for transfer learning). While KnowIT is collected from the TV show \textit{The Big Bang Theory}, KnowIT-X is from another popular TV show: \textit{Friends}. We choose \textit{Friends} because it shares strong similarities with the original KnowIT VQA dataset in terms of format, video length, and audience, whereas it is different in terms of plots and topics, making it ideal for transfer learning. Additionally, to study transfer learning at multiple scales (transfer learning is commonly applied to small-scale datasets), we create three different versions: KnowIT-X full, KnowIT-X v-5k, and KnowIT-X v-3k, with all the collected samples, $5,092$ samples, and $2,900$ samples, respectively. Details are shown in Table \ref{tab:dataset}. Some examples of KnowIT-X are shown in the supplementary material.

\begin{table}[t]
\small
\centering
\caption{Comparison between original KnowIT and KnowIT-X datasets. The three versions (v) of KnowIT-X are designed to study transfer learning on knowledge-based VideoQA.}
\vspace{5pt}
\begin{tabular}{lrrrr}
\hline
 &  & \multicolumn{3}{c}{KnowIT-X} \\ \cline{3-5}
 & KnowIT~\cite{garcia2020knowit} & v-3k & v-5k & v-Full \\ \hline \hline
Num.~Episodes & 207 & 25 & 50 & 202 \\
Num.~Scenes & 2,472 & 311 & 613 & 2,565 \\
Num.~Clips & 12,078 & 1,535 & 2,989 & 12,176 \\
Num.~Samples & 24,282 & 2,900 & 5,092 & 21, 412 \\
Ave.~Length of Questions & 7.49 & 7.69 & 7.83 & 7.65 \\
Ave.~Length of Substitles & 56.79 & 38.48 & 39.06 & 39.06 \\
Ave.~Length of Wrong Answers & 4.13 & 1.99 & 2.10 & 2.06 \\
Ave.~Length of Correct Answers & 4.54 & 2.21 & 2.29 & 2.30 \\
Ave.~Length of Knowledge & 10.39 & 14.37 & 15.00 & 14.44 \\ \hline
\end{tabular}
\label{tab:dataset}
\end{table}

\vspace{-15pt}
\paragraph{Video clip extraction}
We obtain videos and subtitles from the original DVDs. We use $202$ episodes, distributed in all $10$ seasons of 
\textit{Friends}. We divide each episode into $20$-second clips by scene, where each scene is aligned with episode transcripts available online.\footnote{\url{https://fangj.github.io/friends/}} We acquire $12,176$ clips in total.

\vspace{-14pt}
\paragraph{Annotations}
Following KnowIT, for each video clip, we ask workers on Amazon Mechanical Turk\footnote{\url{https://www.mturk.com}} to write a question, four candidate answers (one correct), and the associated knowledge that makes the answer correct (see examples in Figure \ref{fig:overview}). After two sweeps of all video clips, we obtain $21,412$ samples, which we split into $17,583$ for training, $1,748$ for validation, and $2,081$ for test sets. For the smaller versions v-5k and v-3k, we randomly reduce the training set to $4,269$ samples and $2,432$ samples, respectively. The same reduction is also applied to the test set and the validation set. The annotation process took us 2 months.

\begin{figure}
\centerline{\includegraphics[width=0.9\linewidth]{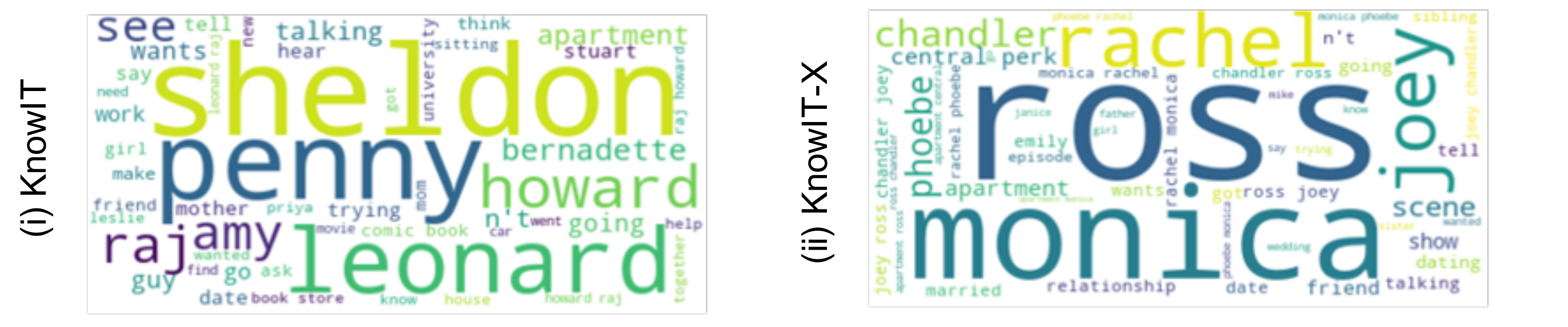}}
\vspace{10pt}
\caption{Vocabulary clouds for KnowIT and KnowIT-X VQA datasets.}
\label{fig:vocab-diff}
\end{figure}

\vspace{-14pt}
\paragraph{Domain comparison} While KnowIT is from the show \textit{The Big Bang Theory} with scientific topics, KnowIT-X is focused more on the relationships among characters. This difference can significantly influence the audience's impressions and interests, causing a distribution gap between the annotations. To show this gap, we first compare the vocabularies. As illustrated in Figure \ref{fig:vocab-diff}, the vocabularies differ a lot. Additionally, we present the distributions of question types in Figure \ref{fig.q_type}, where the type is defined as the first word of each question. With more ‘who’ questions, KnowIT-X is more focused on the vision than KnowIT, which is predominately focused on knowledge, with a larger amount of ‘why’ questions. More comparisons between KnowIT and KnowIT-X are presented in the supplementary material.

\section{Experiments}

We use PyTorch~\cite{NEURIPS2019_9015} to implement our models. The hyperparameters are set following \cite{garcia2020knowit}. In the DET module, the specific entity names include $18$ different entity types.\footnote{Detailed entity types can be found in \url{https://spacy.io/models/en}} 
The DA module employs German as the pivot language and $\alpha = 0.998$. 

\vspace{-14pt}
\paragraph{Evaluation metrics} Following \cite{garcia2020knowit}, we report results both on the knowledge retrieval and the video reasoning modules. Knowledge retrieval results are reported as recall at $k$ (R@$k$) with $k = 1, 5, 10$, and median rank (MR). Video reasoning results are reported in terms of accuracy. For the knowledge retrieval, we use MR as our primary metric and R@5 as the secondary metric, as the top-5 retrieved sentences are the ones used in the video reasoning module.  Additionally, note that as MR and recall do not consider the potential similarities between other highly related samples in the knowledge base, the evaluation of the knowledge retrieval may not be representative of the real improvement on the final VideoQA task. 

\vspace{-14pt}
\paragraph{Domain-agnostic knowledge transfer with DET} First, we evaluate the domain-agnostic transfer learning on the knowledge retrieval module.   Table \ref{tab:tf-knowitx} shows results on KnowIT-X v-Full. We compare different learning methods: direct learning on a single dataset (KnowIT, KnowIT-X, or both), and transfer learning with and without our proposed DET. Direct training on the original KnowIT (row 1) shows the worst performance by a large margin, confirming the big domain gap between the two datasets. Transfer learning without DET (row 4), which is equivalent to standard finetuning, does not improve results over direct training (row 2), verifying that in this case the model is not able to discern between domain-specific and domain-agnostic knowledge. However, when we incorporate our proposed DET (row 5), results are successfully improved, obtaining the best R@5 and MR. Direct training on both datasets (row 3) performs similar to our method, but it requires more time and memory.\footnote{This is because the knowledge retrieval module behaves as $O(N^2)$, where $N$ is the number of training samples.} 

\begin{table}[t]
\small
\caption{Domain-agnostic knowledge transfer results on the KnowIT-X v-Full.}
\vspace{-5pt}
\begin{center}
\begin{tabular}{c l l l c c c r }
\hline
& Source & Target & Learning & R@1 &	R@5 &	R@10 & MR \\ \hline
\hline
1 & KnowIT & - & Direct &	0.178&	0.363&	0.444&	17 \\
2 & KnowIT-X & - & Direct &	0.303&	0.528&	0.618&	5\\
3 & Both &	- &	Direct & 0.316&	0.531&	0.621&	4\\
4 & KnowIT &	KnowIT-X & Transfer (w/o DET) &	0.299&	0.529&	0.614&	5 \\
5 & KnowIT & KnowIT-X & Transfer (w/ DET) & 0.303 & 0.531 &	0.630 &	4 \\
\hline
\end{tabular}
\end{center}
\label{tab:tf-knowitx}
\end{table}

\begin{table}[t]
\caption{Knowledge retrieval module results on KnowIT-X v-3k and v-5k. Each training configuration is computed with and without DET and DA.}
\vspace{5pt}
\centering
\small
\begin{tabular}{lllrrcrr}
\hline
& &  &  \multicolumn{2}{c}{v-3k} & & \multicolumn{2}{c}{v-5k} \\ \cline{4-5} \cline{7-8}
Source & Target & Learning & R@5 & MR & & R@5 & MR \\ \hline \hline
KnowIT & - & Direct & 0.010 & 7,864 & & 0.010 & 7,798 \\
\rowcolor[HTML]{EFEFEF}
 &  & \quad w/ DET+DA & 0.421 & 13 & &  0.397 & 18\\ 
KnowIT-X & - & Direct & 0.472 & 8 & & 0.490 & 6 \\
\rowcolor[HTML]{EFEFEF}
 &  &  \quad w/ DET+DA &  0.497 &  7 &  &  0.490 &  6 \\ 
Both & - & Direct & 0.508 & \textbf{5} & & 0.524 & 5 \\
\rowcolor[HTML]{EFEFEF}
 &  &  \quad w/ DET+DA &  0.487 &  6 &  &  0.497 &  6 \\ 
KnowIT & KnowIT-X & Transfer & 0.482 & 6 & & 0.526 & 5 \\
\rowcolor[HTML]{EFEFEF}
 &  &  \quad w/ DET+DA &  \textbf{0.521} &  \textbf{5} &  &  \textbf{0.537} &  \textbf{4} \\ 
\hline
\end{tabular}
\label{tab:5k-bk-NER}
\end{table}

\vspace{-14pt}
\paragraph{Small-scale knowledge transfer} The benefits of using the proposed transfer learning are even more obvious on the smaller versions of the dataset. Results on KnowIT-X v-3k and v-5k are reported in Table \ref{tab:5k-bk-NER}.  We apply DET and DA to all the training configurations, obtaining consistent improvements and indicating that the proposed method captures the shared knowledge between different datasets. The only exception is when we train directly on both datasets (KnowIT and KnowIt-X), as the model can obtain enough information at training time at the expense of a higher computation cost. Even so, our proposed transfer learning method obtains the best performance both in KnowIT-X v-3k and KnowIT-X v-5k. Results with other metrics are presented in the supplementary material.

\begin{table}
	\begin{minipage}{0.47\linewidth}
		\centering
		\small
		\vspace{25pt}
  \setlength{\tabcolsep}{3pt}
  \begin{tabular}{ l c r r r r }
    \hline
    DET &  R@1 & R@5 & R@10 & MR \\\hline \hline
    Mask-out &     0.241&	0.442&	0.534&	8\\
    Hyphen sep.  & 0.294&	0.530&  0.633&	4\\
    Appositive   &  \textbf{0.303}&	\textbf{0.531}&	\textbf{0.630}&	\textbf{4}\\
    \hline
    \end{tabular}
    \vspace{25pt}
    \caption{Knowledge retrieval results using different DET methods. }
	\label{table:DET-analysis}
	\end{minipage}\hfill
	\begin{minipage}{0.47\linewidth}
		\centering
		\includegraphics[width=0.90\linewidth]{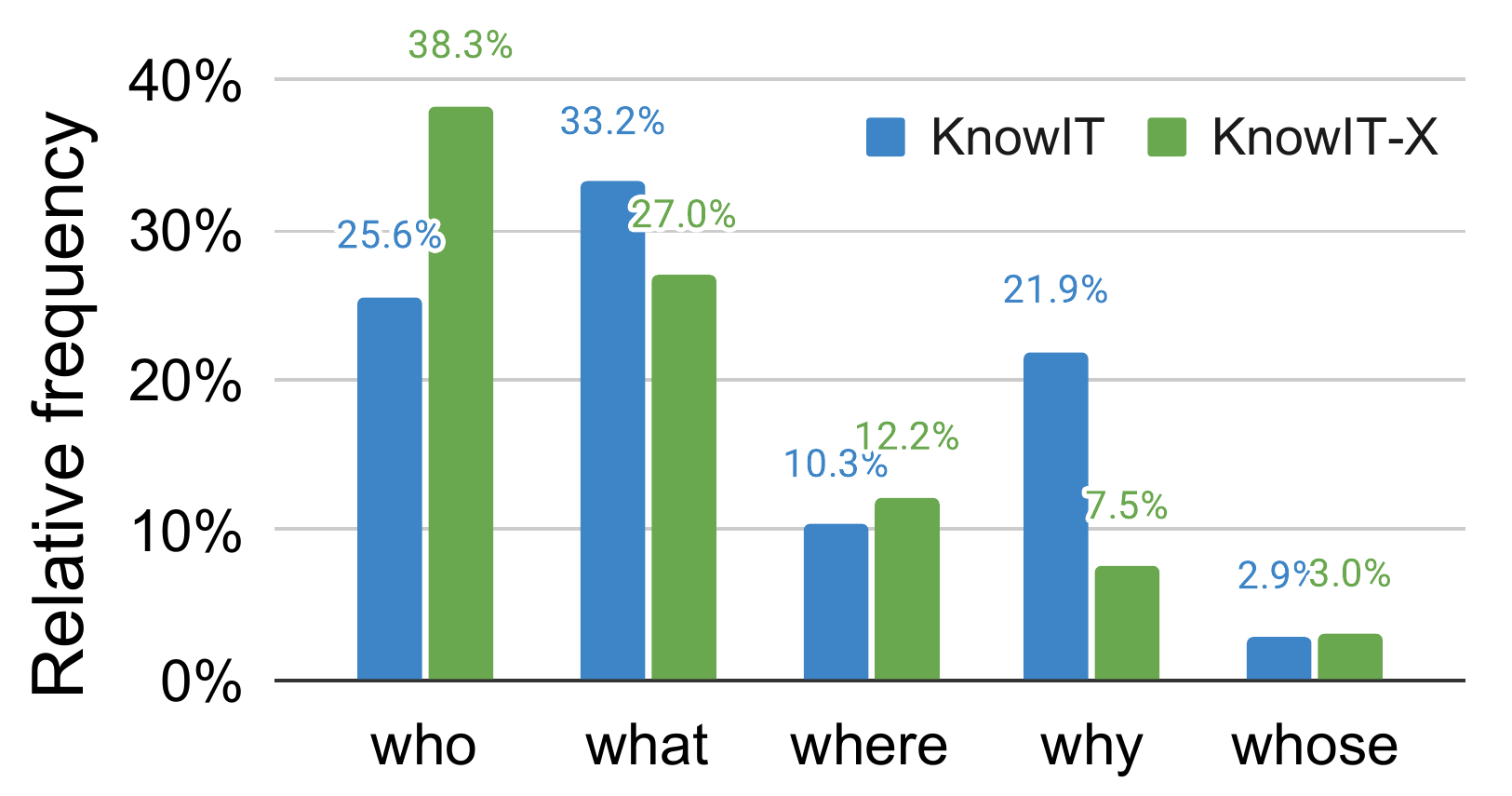}
		\vspace{10pt}
        \captionof{figure}{The distributions of question types in KnowIT and KnowIT-X VQA dataset.}
		\label{fig.q_type}
	\end{minipage}
\end{table}  

\begin{table}[t]
\small
    \centering
    \caption{Video reasoning accuracy with direct learning (w/o transfer learning) on KnowIT and KnowIT-X datasets, using different combination of inputs (vision, language, and knowledge). Image features are used as the vision source in this experiment.}
    \vspace{5pt}
    \begin{tabular}{ccllcccc}
         \hline
         & & &  &  & \multicolumn{3}{c}{KnowIT-X} \\ \cline{6-8}
 & Vision & Language & Knowledge & KnowIT~\cite{garcia2020knowit} & v-3k & v-5k & v-Full \\ \hline \hline
         1 & - & QA & - & 0.530	& 0.376	& 0.333 & 0.482 \\
         2 & - & QA, Subs & - &	0.587 &	0.497 &0.515 & 0.622 \\
         3 &  \cmark & {QA, Subs} & {-} &	{0.587} & {0.497} & {0.516} & {0.623} \\
         4 & \cmark  & QA, Subs & Retrieved &  0.652	& 0.579 & 0.663 &	0.738\\
         5 & - & QA, Subs & GT & 0.731 & 0.720 & 0.787 & 0.845 \\
         \hline
    \end{tabular}
    \label{tab:baseline}
\end{table}

\begin{table}[]
\centering
\caption{Video reasoning accuracy with the proposed transfer learning methods on KnowIT-X v-3k, v-5k and v-full. Transfer is pre-trained on KnowIT.}
\small
\vspace{5pt}
\begin{tabular}{lllllllr}
\hline
{} &  &  &  &  & \multicolumn{3}{c}{KnowIT-X} \\ \cline{6-8} 
{Vision} & {Learning} & {Knowledge} & {DET} & {DA} & {v-3k} & {v-5k} & {v-full} \\ 
\hline
\hline
{Image} & {Direct} & {Retrieved} & {-} & {-} & {0.579} & {0.663} & 0.738 \\
{Image} & {Transfer} & {Retrieved} & {-} & {-} & {0.584} & {0.678} & \multicolumn{1}{l}{{ 0.721}} \\
{Image} & {Transfer} & {Retrieved} & {-} & {\cmark} & {0.629} & {0.698} & \multicolumn{1}{l}{{ 0.731}} \\
{Image} & {Transfer} & {Retrieved} & {\cmark} & {-} & {0.660} & 0.692 & {{ 0.726}} \\
{Image} & {Transfer} & {Retrieved} & {\cmark} & {\cmark} & {0.665} & {0.717} & \multicolumn{1}{l}{{ 0.740}} \\ \hline
{Caption} & {Direct} & {Retrieved} & {-} & {-} & {0.624} & {0.669} & \multicolumn{1}{l}{{ 0.756}} \\
{Caption} & {Transfer} & {Retrieved} & {-} & {-} & {{0.629}} & {{0.675}} & { 0.734} \\
{Caption} & {Transfer} & {Retrieved} & {-} & {\cmark} & {{0.645}} & {{0.701}} & { 0.747} \\
{Caption} & {Transfer} & {Retrieved} & {\cmark} & {-} & {{0.670}} & {{0.712}} & { 0.739} \\
{Caption} & {Transfer} & {Retrieved} & {\cmark} & {\cmark} & {{0.690}} & {{0.723}} & { 0.758} \\ \hline
{Facial} & {Direct} & {Retrieved} & {-} & {-} & {0.579} & {0.663} & { 0.739} \\
{Facial} & {Transfer} & {Retrieved} & {-} & {-} & {{0.584}} & {{0.678}} & { 0.721} \\
{Facial} & {Transfer} & {Retrieved} & {-} & {\cmark} & {{0.629}} & {{0.698}} & { 0.731} \\
{Facial} & {Transfer} & {Retrieved} & {\ \cmark} & {\ -} & {0.660} & {0.692} & {0.726} \\
{Facial} & {Transfer} & {Retrieved} & {\cmark} & {\cmark} & {{0.665}} & {{0.717}} & {0.740} \\
\hline
{-} & {-} & {GT} & {\cmark} & {\cmark} & {0.802} & {0.821} & { 0.850}\\
\hline
\end{tabular}
\label{tab:full-ver-vr-result}
\end{table}
  
\vspace{-14pt}
\paragraph{DET analysis}

We compare different ways of inserting the recognised named entities with DET and show that the appositive tagging is better than other tagging methods. In particular, we evaluate \textit{Mask-out}, in which the detected entity is removed and replaced by its label (e.g. \textit{Why was \underline{person} acting weird?}), and \textit{Hyphen separator}, in which the entity and its label are separated by the character ``-" (e.g. \textit{Why was \underline{Chandler-person}, acting weird?}).  Results are reported Table  \ref{table:DET-analysis} for the model pre-trained on KnowIT and transferred to KnowIT-X v-Full.

\vspace{-14pt}
\paragraph{Video reasoning direct results}
We evaluate the standard ROCK model~\cite{garcia2020knowit} (\ie, without transfer learning) on the three versions of KnowIT-X and compare the results with the original KnowIT dataset. We use image features for the vision; QA, and subtitles (Subs) for the language; and retrieved and ground truth knowledge for the KB. All the models are trained from scratch. Accuracy is shown in Table \ref{tab:baseline}. We observe that 1) the accuracy when only using questions and answers (row 1) is lower on KnowIT-X, indicating that the language bias is not as strong in KnowIT-X, 2) the small-scale datasets perform significantly worse than its full counterpart, indicating the importance of training data. 3) the performances with retrieved knowledge (row 4) surpass the one without knowledge (row 3) by a large margin, which demonstrates that the retrieved knowledge predicts the correct answers.

\begin{figure}
    \centering
    \includegraphics[width=0.90\linewidth]{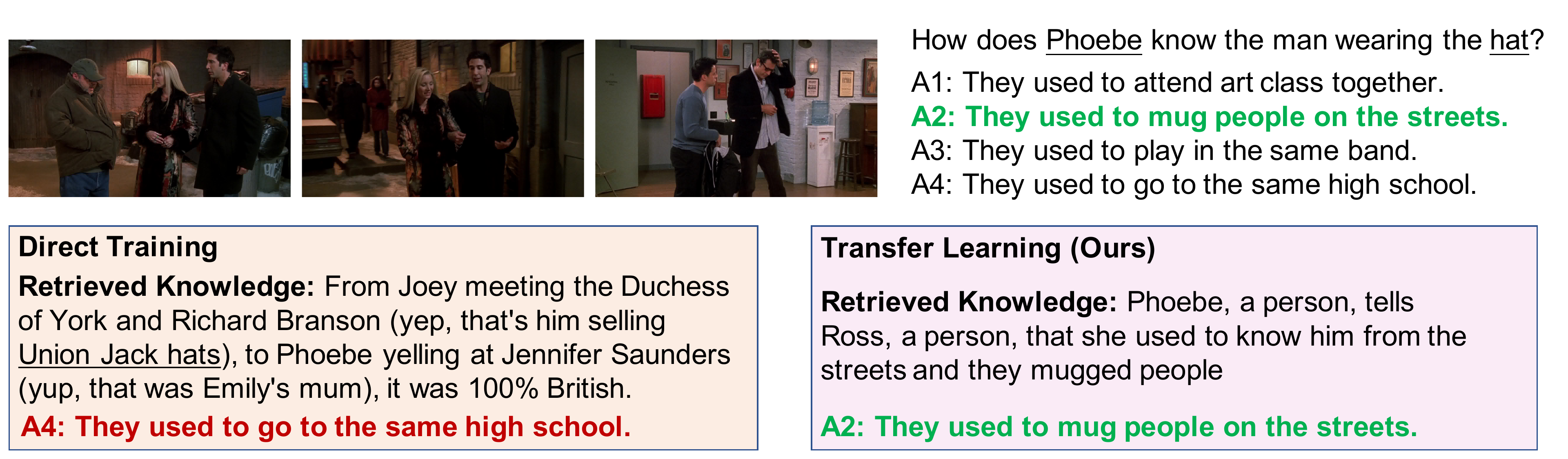}
    \vspace{5pt}
    \caption{Example results: (left) direct learning, (right) proposed transfer learning.}
\label{fig:visual-example}
\end{figure}

\vspace{-14pt}
\paragraph{Video reasoning transfer results}
In Table \ref{tab:full-ver-vr-result}, we show the video reasoning results on the KnowIT-X v-3k, v-5k and v-full with different scales of visual information. We compare the full transfer learning method against variations with and without DET and DA, and the direct learning method. It can be seen that each strategy contributes to improving the final VideoQA accuracy. When both DET and DA are included, the result is the best. Also, among all the vision sources, captions usually help acquire the best performance. Additionally,  in Figure \ref{fig:visual-example}, we show a qualitative comparison between direct and transfer learning. In direct learning, the retrieved knowledge is too much focused on the domain-specific entity `hat' and has limited contributions to the final answer. On the contrary, transfer learning successfully recognises it as a domain-specific entity, putting more emphasis on the relationship between people. The retrieved knowledge is highly connected with the question and the correct answer.

\section{Conclusion}
We proposed a knowledge-based transfer learning framework for VideoQA. We divided the knowledge learned by the VideoQA models into two categories and achieved better transfer results by mitigating the influence of the domain-specific knowledge and transferring the domain-agnostic knowledge. Back translation was applied to augment small-scale datasets. By constructing a new knowledge-based VideoQA dataset, the experimental results showed that domain-agnostic knowledge can be transferred adaptively from the source dataset and boost performance in the proposed framework. 
\vspace{-14pt}
\paragraph{Acknowledgements}
This work was supported by JSPS
KAKENHI No. JP18H03264 and JP20K19822, and
JST ACT-I.

\bibliography{egbib}
\end{document}


\maketitle

This supplementary material contains the analysis of KnowIT-X VQA dataset as well as more results for knowledge retrieval on KnowIT-X v-3k and v-5k.

\section{Dataset Statistics}
\subsection{Dataset examples}
\begin{figure}[htbp!]
    \centering
    \includegraphics[width=11.5cm]{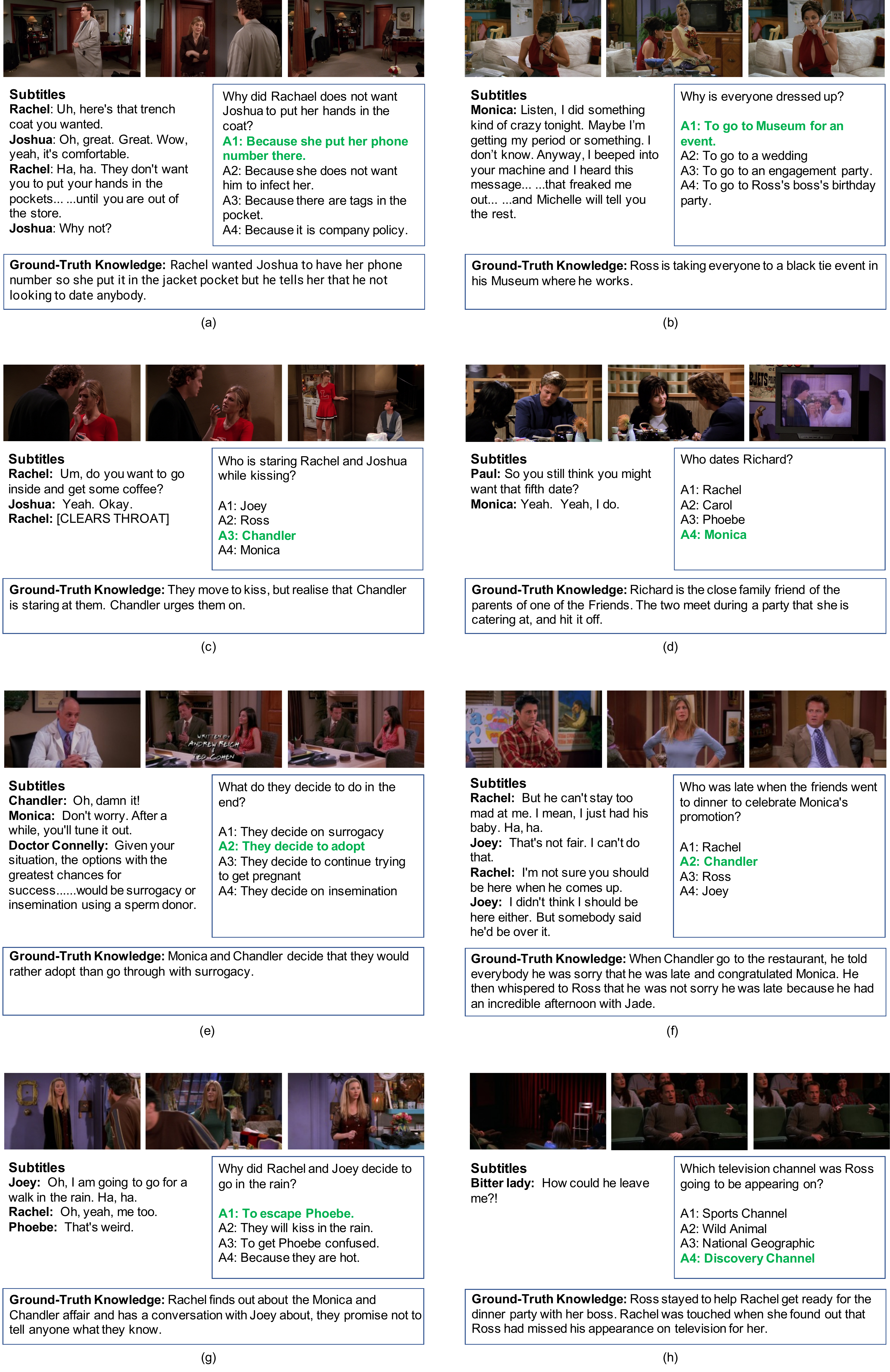}
    \vspace{-4pt}
    \caption{Examples from KnowIT-X VQA Dataset. We show in the figure three frames of the given video clips, QA pairs, subtitles and the annotated ground-truth knowledge. The correct answers are illustrated in green.}
    \label{fig:dataset-example}
\end{figure}

\begin{figure}[htbp!]
    \centering
    \includegraphics[width=11.5cm]{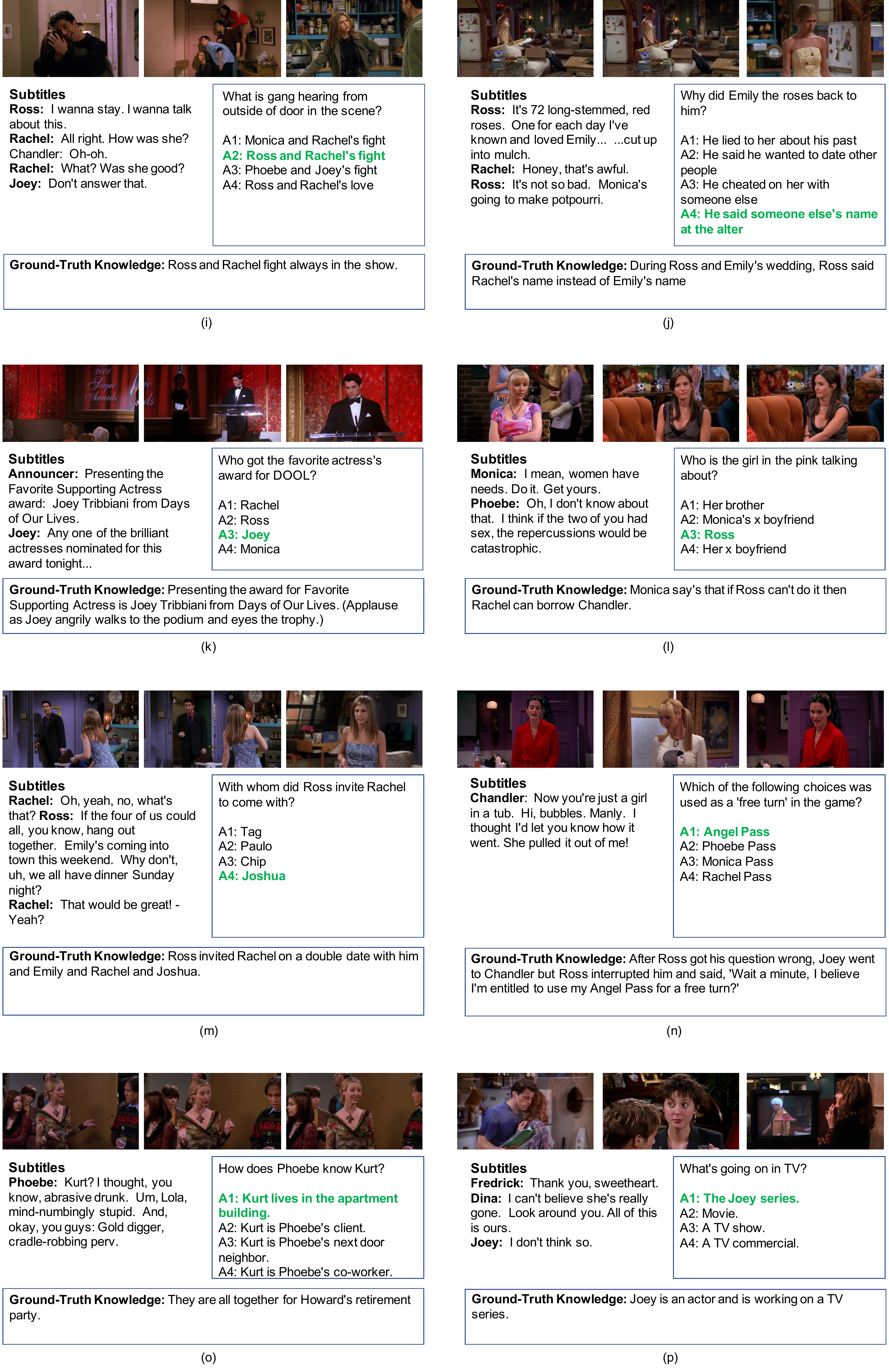}
    \vspace{-4pt}
    \caption{More examples from KnowIT-X VQA Dataset. }
    \label{fig:dataset-example-2}
\end{figure}

In the main paper, we constructed a new knowledge-based dataset KnowIT-X VQA, of which each sample includes a video clip, a related question, four candidate answers, subtitles and a piece of annotated ground-truth knowledge. We give some examples of the dataset in Figure~\ref{fig:dataset-example}  and Figure \ref{fig:dataset-example-2}.

\subsection{Dataset comparisons of the vocabulary}
\begin{figure}[t]
\begin{minipage}{.45\linewidth}
    \centerline{\includegraphics[width=6.0cm]{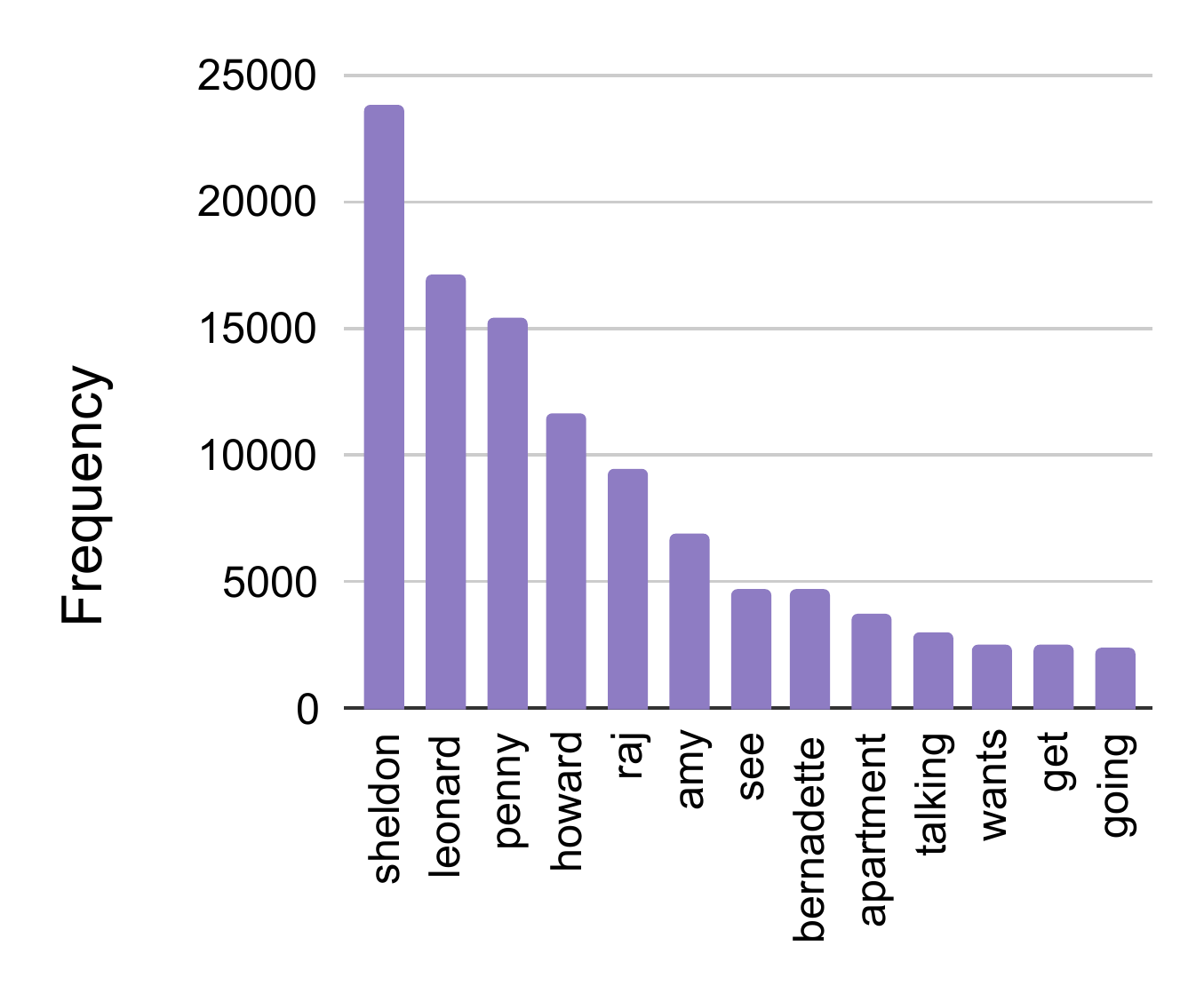}}
    \centerline{(a) Vocabulary distribution in KnowIT.}
\end{minipage}
\hfill
\begin{minipage}{.45\linewidth}
    \centerline{\includegraphics[width=6.0cm]{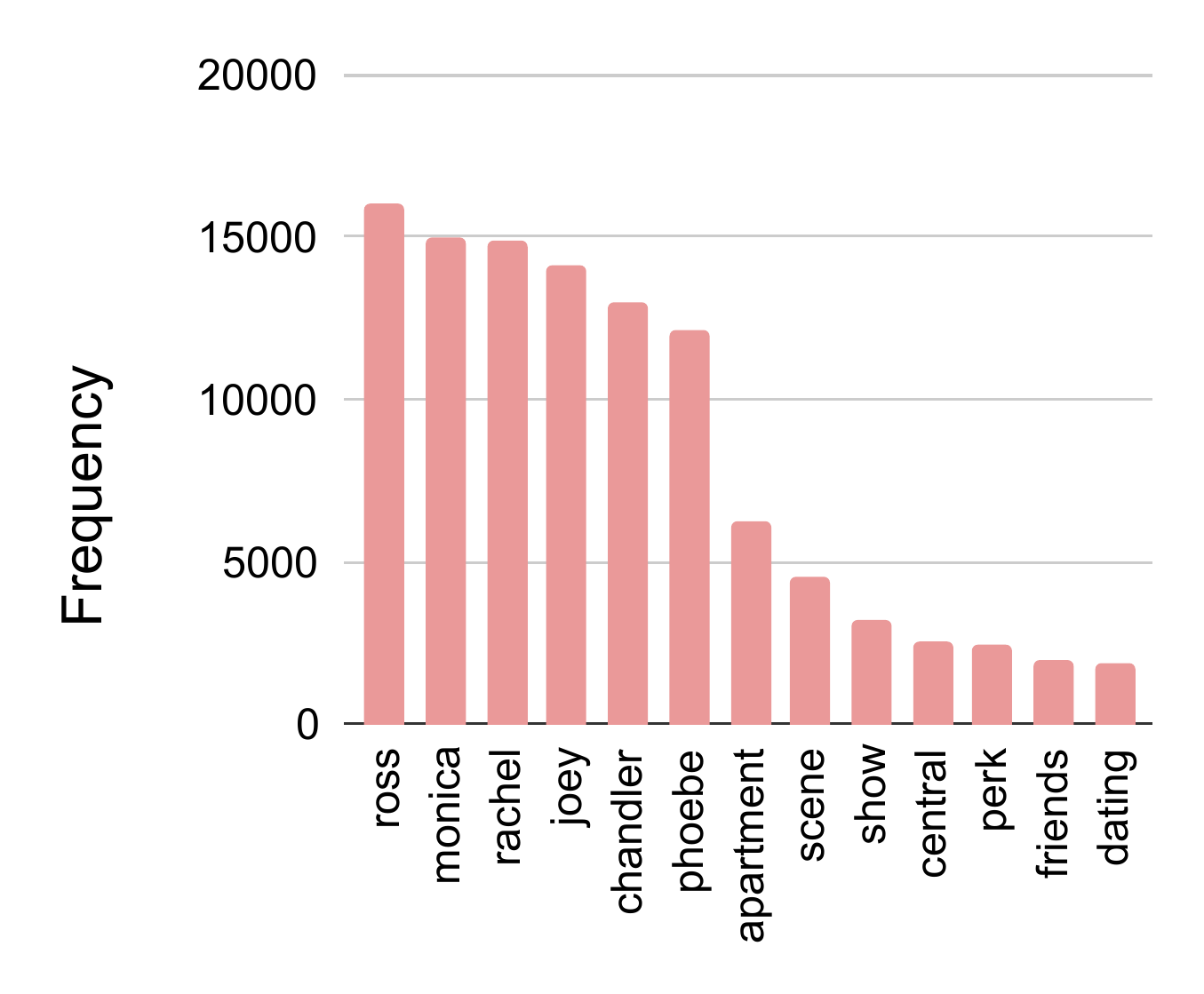}}
    \centerline{(b) Vocabulary distribution in KnowIT-X.}
\end{minipage}
\vspace{4pt}
\caption{Vocabulary distribution in the two datasets, where x-axis shows words in the vocabulary, and y-axis their frequency in the questions, answers and annotated knowledge.}\label{fig:vocab-dist}
\end{figure}

We compare KnowIT-X VQA for transfer knowledge with the source dataset KnowIT VQA. As shown in the word clouds in the Figure 3 of our main paper, the vocabularies in KnowIT and KnowIT-X differ a lot. As an extension for the figure, the vocabulary distributions of the two datasets are illustrated in Figure \ref{fig:vocab-dist}. Before calculating the distributions, we first remove the stopwords, which refer to the words that do not carry any information such as prepositions, conjunctions, etc. 
We use the stopwords list for English from Natural Language Toolkit (NLTK) and add several words like ‘n't’ and ‘ah’. It can be seen in Figure \ref{fig:vocab-dist} that the most frequent words are almost domain-specific entities like character names and locations.

\section{More Experiment Results}
Table \ref{tab:extend-fig3} is the extension of Table 3 in the main paper. R@$1$ and R@$10$ are shown in the table. Although not all the experiments with the proposed transfer learning method show the best result, it is the R@$5$ that contributes most to the video reasoning as only the first five most relevant pieces of knowledge are used as inputs.

\vspace{-100pt}
\begin{table}[t!]
\caption{More knowledge retrieval module results on KnowIT-X v-3k and v-5k. Each training configuration is computed with and without DET and DA.}
\centering
\small
\begin{tabular}{lllrrcrr}
\hline
& &  &  \multicolumn{2}{c}{v-3k} & & \multicolumn{2}{c}{v-5k} \\ \cline{4-5} \cline{7-8}
Source & Target & Learning & R@1 & R@10 &  & R@1 &R@10  \\ \hline \hline
KnowIT & - & Direct & 0.005 & 0.010 & & 0.005 & 0.010 \\
\rowcolor[HTML]{ECF4FF}
 &  & \quad w/ DET+DA &  0.244&0.482 & &  0.190&0.447\\ 
KnowIT-X & - & Direct & 0.242 & 0.523 & & 0.304 & 0.598 \\
\rowcolor[HTML]{ECF4FF}
 &  &  \quad w/ DET+DA &  0.279 & 0.563 &  &  \textbf{0.312} &  0.592 \\ 
Both & - & Direct & 0.299 & 0.569 & & 0.278 & \textbf{0.612} \\
\rowcolor[HTML]{ECF4FF}
 &  &  \quad w/ DET+DA &  0.239 &  0.574 &  &  0.277 &  0.608 \\ 
KnowIT & KnowIT-X & Transfer & 0.294 & \textbf{0.589} & & 0.270 & 0.592 \\
\rowcolor[HTML]{ECF4FF}
 &  &  \quad w/ DET+DA &  \textbf{0.299} &0.579  &&0.302 &  \textbf{0.612} \\ 
\hline
\end{tabular}
\label{tab:extend-fig3}
\end{table}